\documentclass{article}
\usepackage{spconf,amsmath,graphicx,hyperref}
\usepackage{tikz}
\usepackage{float}
\usepackage{amssymb}
\usepackage{bm}
\usepackage{algorithm}
\usepackage{algorithmic}
\usepackage[numbers,sort&compress]{natbib}

\newcommand{\ba}{\bm{a}}

\newcommand{\bG}{\bm{G}}

\newcommand{\bx}{\bm{x}}

\newcommand{\bA}{\bm{A}}
\newcommand{\bB}{\bm{B}}

\newcommand{\bb}{\bm{b}}

\newcommand{\bW}{\bm{W}}

\newcommand{\mL}{\mathcal{L}}

\title{Automatic Rank Allocation for Low-Rank Adaptation in Large Language Models via $\ell_p$ Regularization}
\name{
Zebang Xie$^{1}$,
Chuanyang Zheng$^{2}$,
Yik-Chung Wu$^{3}$,
Yihang Gao$^{4,*}$\thanks{$^{*}$Corresponding author: Yihang Gao
(\texttt{gaoyihang@hkbu.edu.hk}).}
}

\address{
$^{1}$National University of Singapore, Singapore \hspace{1em}
$^{2}$Huawei, Hong Kong\\
$^{3}$The University of Hong Kong, Hong Kong\hspace{1em}
$^{4}$Hong Kong Baptist University, Hong Kong\\
}

\begin{document}
%
\maketitle
\begin{abstract}
Low-rank adaptation (LoRA) has become a popular parameter-efficient fine-tuning method for large language models. A key challenge in LoRA is how to determine the rank of each adaptation matrix, as rank directly controls its capacity and efficiency. Existing adaptive-rank methods typically allocate ranks according to manually designed importance scores, which are not directly derived from an optimization objective. In this work, we propose $\ell_p$-LoRA, a principled rank-allocation method based on $\ell_p$ regularization with $0<p<1$, which is a classical sparsity-inducing technique in signal processing and statistics. Specifically, we regularize the energy of each rank-one LoRA component, encouraging redundant components to vanish while preserving important ones. We derive the corresponding proximal subproblem and reduce the matrix optimization to a two-dimensional problem, leading to an implicit thresholding criterion for identifying redundant components. Experiments on natural language understanding and question-answering tasks demonstrate that the proposed method achieves competitive performance with existing LoRA baselines.

\end{abstract}

\vspace{-0.5em}
\section{Introduction}
\label{sec:intro}
\vspace{-0.5em}
Large language models (LLMs) have achieved remarkable performance across a wide range of downstream tasks~\cite{brown2020language}. However, as model sizes continue to grow, fully fine-tuning all model parameters becomes increasingly expensive in both computation and memory. Parameter-efficient fine-tuning (PEFT) has therefore emerged as an effective approach for adapting pretrained models to downstream tasks with substantially fewer trainable parameters~\cite{houlsby2019parameter}. Among various PEFT methods, low-rank adaptation (LoRA)~\cite{hu2022lora} has become one of the most widely adopted approaches. LoRA freezes the pretrained weights and represents each trainable weight update by a low-rank factorization, thereby significantly reducing the number of trainable parameters.

An important question in LoRA is how to determine the rank of each adaptation matrix~\cite{zhang2023adaptive}. Standard LoRA typically assigns a pre-specified rank uniformly across different layers and modules. However, different components of a pretrained model may require different adaptation capacities for a given downstream task, making a uniform rank allocation potentially inefficient. From a resource-allocation perspective, given a total rank budget determined by computational or memory constraints, the problem is therefore how to distribute the available ranks across different LoRA matrices.

Existing adaptive-rank methods have primarily addressed this problem by designing importance measures for individual LoRA components. The pioneering work AdaLoRA~\cite{zhang2023adaptive}, for example, dynamically allocates the rank budget according to importance scores constructed from information collected during training. Subsequent methods further refine such criteria by incorporating additional gradient or layer-wise information~\cite{cui2026igulora,valipour2023dylora}. These approaches determine rank allocation through designed importance scores rather than deriving the pruning criterion directly from an explicit optimization formulation. Another line of work considers probabilistic or Bayesian formulations for adaptive rank determination~\cite{yang2024bayesian,duan2026bara}. Such approaches provide an alternative perspective, but typically introduce additional modeling and inference mechanisms beyond standard LoRA training.

In this work, we seek a principled, mathematically motivated rank-allocation method for LoRA. We observe that adaptive rank allocation can be naturally interpreted as a structured sparsity problem: each rank-one LoRA component can be viewed as a candidate component to be either retained or removed. Motivated by the sparsity-inducing properties of $\ell_p$ regularization with $0<p<1$, which has been widely studied in signal processing and statistical modeling~\cite{chartrand2007exact}, we formulate rank selection by imposing an $\ell_p$ penalty on the energy of rank-one LoRA components. Under this formulation, redundant components are encouraged to vanish, while informative components tend to be retained.
	
	
	

\vspace{-0.5em}
\section{Background}
\label{sec:format}
\vspace{-0.5em}
\subsection{Low-Rank Adaptation}
Consider LoRA-based fine-tuning~\cite{hu2022lora} for a multi-layer neural network $\phi(\cdot;\Theta)$, such as a transformer-based LLM~\cite{vaswani2017attention}. 
Let $\Theta := \{\bW^{(1)}, \bW^{(2)}, \cdots, \bW^{(L)}\}$ denote the collection of parameter matrices equipped with LoRA. For each parameter matrix, the fine-tuned parameter is represented as
\begin{equation}
	\label{eq_lora}
	\bW_{\text{ft}} = \bW_{\text{pt}} + \Delta\bW := \bW_{\text{pt}} + \bB \bA,
\end{equation}
where $\bW_{\mathrm{pt}}\in\mathbb{R}^{m\times n}$ and $\bW_{\mathrm{ft}}\in\mathbb{R}^{m\times n}$ denote the pretrained and fine-tuned parameter matrices, respectively, and $\Delta\bW\in\mathbb{R}^{m\times n}$ is the corresponding adaptation matrix.
We take the trainable LoRA factors
$\bA\in\mathbb{R}^{r\times n}$ and
$\bB\in\mathbb{R}^{m\times r}$, where $r\ll \min\{m,n\}$. 
For notational simplicity, throughout the paper we present the formulation and derivation for a single LoRA-equipped parameter matrix and suppress the layer or module index. The same construction can be applied to all parameter matrices equipped with LoRA.

The rank $r$ determines the number of rank-one components available for adapting each parameter matrix. Since different layers and modules can exhibit different levels of relevance to a downstream task, assigning an identical rank to all LoRA modules may lead to inefficient use of the available rank budget~\cite{zhang2023adaptive}. Some parameter matrices may be sufficiently adapted using only a few rank-one components, whereas others may require a larger adaptation capacity. At the same time, unnecessarily large ranks increase the number of trainable parameters and may introduce redundant degrees of freedom. These observations motivate adaptive rank allocation, where different parameter matrices are assigned different effective ranks according to their contribution to the downstream task~\cite{valipour2023dylora,cui2026igulora}.

\subsection{$\ell_p$ Regularization}

Sparsity-promoting regularization has been widely studied in signal processing, sparse recovery, and statistical estimation~\cite{chartrand2007exact}. A representative formulation is
\begin{equation}
	\min_{\bx \in \mathbb{R}^d} f(\bx)+\lambda\|\bx\|_{p}^p
	=
	f(\bx)+\lambda\sum_{i=1}^{d}|x_i|^p,
	\quad 0<p\leq 1,
	\label{eq_lp}
\end{equation}
where $f$ denotes the data-fitting objective and $\lambda>0$ controls the regularization parameter. The classical $\ell_1$ penalty provides a convex surrogate for the $\ell_0$ sparsity measure, while the nonconvex $\ell_p$ penalty with $0<p<1$ more closely approximates $\ell_0$ and generally induces stronger sparsity~\cite{candes2006robust,chartrand2007exact}.

A key property of $\ell_p$ regularization is its thresholding behavior. Given a reference point $\tilde{\bx}$, the corresponding proximal subproblem takes the form
\begin{equation}
	\min_{\bx}
	\frac{1}{2}\|\bx-\tilde{\bx}\|_2^2
	+\lambda\|\bx\|_p^p.
	\label{eq_lp_prox}
\end{equation}
For each coordinate, the problem reduces to a scalar minimization in which sufficiently small entries of $\tilde{\bx}$ are mapped exactly to zero, whereas sufficiently large entries are retained as nonzero. This zero--nonzero transition acts as a thresholding rule and provides a natural mechanism for variable selection and sparse recovery.

\section{Method}
\label{sec3}
\subsection{Motivation and Problem Formulation}
As shown in~\eqref{eq_lora}, the trainable parameters in LoRA fine-tuning are the factor matrices ${\bA,\bB}$. Accordingly, the training problem can be written abstractly as
\begin{equation}
	\min_{\bA,\bB} \mL(\bB\bA),
	\label{eq_lora_train}
\end{equation}
where $\mL$ denotes the training loss. Its specific form depends on the underlying model, downstream task, and training data.

The LoRA update admits the rank-one decomposition
\begin{equation}
	\bB\bA
	=
	\sum_{i=1}^{r}
	\bB_{:,i}\bA_{i,:}.
\end{equation}
We define the energy of the $i$-th rank-one component by
\begin{equation}
	e_i
	:=
	\|\bB_{:,i}\bA_{i,:}\|_{\mathrm{F}}
	=
	\|\bB_{:,i}\|_{2}\|\bA_{i,:}\|_{2},
\end{equation}
and collect the component energies into
\begin{equation}
	\Phi(\bA,\bB)
	:=
	[e_1,e_2,\ldots,e_r]^\top
	\in\mathbb{R}^{r}.
	\label{eq_energy}
\end{equation}
A component is active whenever $e_i\neq0$. Hence, $\|\Phi(\bA,\bB)\|_{0}$ counts the number of active rank-one components and provides a natural notion of the effective LoRA rank. In particular, $	\operatorname{rank}(\bB\bA)
\leq
\|\Phi(\bA,\bB)\|_{0}$.
Given a rank budget $b$, rank allocation can therefore be formulated as the cardinality-constrained problem
\begin{equation}
		\min_{\bA,\bB}\quad
		\mL(\bB\bA),\quad
		\mathrm{s.t.}\quad
		\|\Phi(\bA,\bB)\|_0\leq b.
	\label{eq_l0_constraint}
\end{equation}
For multiple LoRA-equipped parameter matrices, $\Phi$ can be understood as the concatenation of their component-energy vectors, and the same formulation applies to a global rank budget.

Problem~\eqref{eq_l0_constraint} is difficult to optimize directly due to the combinatorial and discontinuous property of the $\ell_0$ constraint. A natural alternative is to replace the hard cardinality constraint by a sparsity-promoting penalty. Moreover, for any fixed vector $\bx$, $	\lim_{p\rightarrow0^+}\|\bx\|_p^p=\|\bx\|_{0},$
which motivates the use of the nonconvex $\ell_p$ penalty with $0<p<1$ as a continuous surrogate for the $\ell_0$ measure~\cite{chartrand2007exact}. We therefore consider the following $\ell_p$-regularized LoRA problem:
\begin{equation}
	\min_{\bA,\bB} \quad
	\mL(\bB\bA)
	+
	\lambda
	\|\Phi(\bA,\bB)\|_p^p,
	\label{eq_main_problem}
\end{equation}
or equivalently,
\begin{equation}
	\min_{\bA,\bB}\quad
	\mL(\bB\bA)
	+
	\lambda
	\sum_{i=1}^{r}
	\left(
	\|\bB_{:,i}\|_2
	\|\bA_{i,:}\|_2
	\right)^p,
	\label{eq_main_problem_expanded}
\end{equation}
where $\lambda>0$ is the regularization parameter and $0<p<1$. The resulting penalty acts directly on entire rank-one components rather than individual entries of $\bA$ and $\bB$, thereby promoting sparsity at the rank-component level.

\subsection{Algorithm Design: Implicit Thresholding}

We next develop a tractable procedure based on the $\ell_p$-regularized formulation in~\eqref{eq_main_problem_expanded}. Let $(\bA_k,\bB_k)$ denote the current LoRA factors at iteration $k$. Directly handling the training loss $\mL(\bB\bA)$ together with the nonconvex $\ell_p$ penalty is generally difficult. We therefore construct a local quadratic model of the training loss around $(\bA_k,\bB_k)$.
The resulting local model is
\begin{equation}
	\begin{split}
		\mL(\bB\bA)
		& \approx 
		\mL(\bB_k\bA_k)
		+\left\langle
		\nabla_{\bA}\mL_k,
		\bA-\bA_k
		\right\rangle\\
		&\quad
		+\left\langle
		\nabla_{\bB}\mL_k,
		\bB-\bB_k
		\right\rangle \\
		&\quad+\frac{M}{2}\left\|\bA-\bA_k\right\|_{\mathrm{F}}^2
		+\frac{M}{2}\left\|\bB-\bB_k\right\|_{\mathrm{F}}^2,
	\end{split}
	\label{eq_quad_model}
\end{equation}
where $\nabla_{\bA}\mL_k
:=
\nabla_{\bA}\mL(\bB_k\bA_k),
~
\nabla_{\bB}\mL_k
:=
\nabla_{\bB}\mL(\bB_k\bA_k)$ denote the gradients of the objective with respect to two factors. 
Substituting~\eqref{eq_quad_model} into~\eqref{eq_main_problem_expanded} gives the local regularized subproblem
\begin{equation}
	\begin{aligned}
		\min_{\bA,\bB}
		&
		\left\langle
		\nabla_{\bA}\mL_k,\bA-\bA_k
		\right\rangle
		+
		\left\langle
		\nabla_{\bB}\mL_k,\bB-\bB_k
		\right\rangle 
		+\frac{M}{2}\left\|\bA-\bA_k\right\|_{\mathrm{F}}^2\\
		& +\frac{M}{2}\left\|\bB-\bB_k\right\|_{\mathrm{F}}^2
		+
		\lambda
		\sum_{i=1}^{r}
		\left(
		\|\bB_{:,i}\|_2
		\|\bA_{i,:}\|_2
		\right)^p.
	\end{aligned}
	\label{eq_local_subproblem}
\end{equation}

Define the nominal gradient steps
\begin{equation}
	\widetilde{\bA}
	:=
	\bA_k-\frac{1}{M}\nabla_{\bA}\mL_k,
	\qquad
	\widetilde{\bB}
	:=
	\bB_k-\frac{1}{M}\nabla_{\bB}\mL_k.
	\label{eq_nominal_points}
\end{equation}
By completing the square and removing terms independent of $(\bA,\bB)$, problem~\eqref{eq_local_subproblem} is equivalently written as
\begin{equation}
	\min_{\bA,\bB}
	\frac{M}{2}
	\left(
	\left\|\bA-\widetilde{\bA}\right\|_{\mathrm{F}}^2
	+
	\left\|\bB-\widetilde{\bB}\right\|_{\mathrm{F}}^2
	\right)
	+
	\lambda
	\sum_{i=1}^{r}
	\left(
	\|\bB_{:,i}\|_2
	\|\bA_{i,:}\|_2
	\right)^p.
	\label{eq_prox_subproblem}
\end{equation}
Different from the classical $\ell_p$ proximal problem in~\eqref{eq_lp_prox}, the regularization term in~\eqref{eq_prox_subproblem} is imposed on the energy of each rank-one component, $\|\bB_{:,i}\|_2\|\bA_{i,:}\|_2$. Consequently, the two LoRA factors are coupled within each component, and the standard coordinate-wise thresholding rules for $\ell_p$ regularization cannot be directly applied. Nevertheless, we show below that this coupled problem still admits a tractable reduction and gives rise to an implicit thresholding rule for rank-one LoRA components.

The key observation is that problem~\eqref{eq_prox_subproblem} is separable across the rank-one components. For each $i\in\{1,\ldots,r\}$, define $\ba_i:=\bA_{i,:},~\bb_i:=\bB_{:,i},$
and similarly $\widetilde{\ba}_{i}:=\widetilde{\bA}_{i,:},~\widetilde{\bb}_{i}:=\widetilde{\bB}_{:,i}.$
Then~\eqref{eq_prox_subproblem} can be decomposed into $r$ independent subproblems of the form
\begin{equation}
	\min_{\ba_i,\bb_i}
	\frac{M}{2}
	\left(
	\left\|\ba_i-\widetilde{\ba}_{i}\right\|_2^2
	+
	\left\|\bb_i-\widetilde{\bb}_{i}\right\|_2^2
	\right)
	+
	\lambda
	\left(
	\left\|\ba_i\right\|_2\left\|\bb_i\right\|_2
	\right)^p.
	\label{eq_component_problem}
\end{equation}

Solving~\eqref{eq_component_problem} directly is still challenging, we observe that its dimensionality can be further substantially reduced and the resulting two-dimensional problem becomes solvable.
 In particular, for any fixed magnitudes $\|\ba_i\|_2$ and $\|\bb_i\|_2$, the quadratic terms are minimized when $\ba_i$ and $\bb_i$ are aligned with their corresponding nominal vectors. Hence, an optimal solution must be written as
\begin{equation}
	\ba_i=\alpha_i\widetilde{\ba}_{i},
	\quad
	\bb_i=\beta_i\widetilde{\bb}_{i},
	\quad
	0\leq\alpha_i,\beta_i\leq1.
	\label{eq_scaling}
\end{equation}
Let $s_i:=\|\widetilde{\ba}_{i}\|_2,~t_i:=\|\widetilde{\bb}_{i}\|_2.$
Substituting~\eqref{eq_scaling} into~\eqref{eq_component_problem} reduces the vector optimization problem to the following two-dimensional problem:
\begin{equation}
	\min_{0\leq\alpha,\beta\leq1}
	f_i(\alpha,\beta),
	\label{eq_2d_problem}
\end{equation}
where
\begin{equation}
	f_i(\alpha,\beta)
	:=
	\frac{M}{2}
	\left[
	s_i^2(\alpha-1)^2
	+
	t_i^2(\beta-1)^2
	\right]
	+
	\lambda
	(s_it_i\alpha\beta)^p .
	\label{eq_fi}
\end{equation}
Therefore, the thresholding decision with $(\alpha_i,\beta_i)$ for each rank-one component is obtained by solving the t wo-dimensional problem~\eqref{eq_2d_problem}.

We next briefly characterize the solution of~\eqref{eq_2d_problem}. For a nonzero stationary point with $\alpha,\beta>0$, the first-order optimality conditions are
\begin{equation}
	\begin{split}
	M(\alpha-1)s_i^2 + p\lambda (\beta s_i t_i)^p \alpha^{p-1} = 0, \\
	M(\beta-1)t_i^2 + p\lambda (\alpha s_i t_i)^p \beta^{p-1} = 0.
	\end{split}
	\label{eq_stationary}
\end{equation}
Eliminating one variable from the above system gives
\begin{equation}
	\beta
	=
	\frac{
		1\pm
		\sqrt{
			1-4\alpha(1-\alpha)s_i^2/t_i^2
		}
	}{2}.
	\label{eq_beta_alpha}
\end{equation}
Substituting~\eqref{eq_beta_alpha} back into either first-order condition reduces the original two-variable system to a one-dimensional root-finding problem in $\alpha$, which can be efficiently solved using standard numerical methods, such as the bisection method and Newton's method.

The above characterization only covers nonzero stationary points. Since the $\ell_p$ penalty with $0<p<1$ is non-differentiable when $\alpha\beta=0$, we additionally evaluate the corresponding boundary solutions. Let $\mathcal{S}_i$ denote the set of nonzero stationary solutions obtained from \eqref{eq_stationary} and $\mathcal{B}=\{(0,1),(1,0)\}$ the set of boundary candidates. The solution of~\eqref{eq_2d_problem} is then obtained among all candidate stationary and boundary points:
\begin{equation}
	(\alpha_i,\beta_i)
	\in
	\arg\min_{(\alpha,\beta)\in
		\mathcal{S}_i\cup\mathcal{B}}
	f_i(\alpha,\beta).
	\label{eq_threshold_solution}
\end{equation}

In contrast to the classical $\ell_p$ proximal problem in~\eqref{eq_lp_prox}, our thresholding decision is obtained implicitly by solving the two-dimensional problem~\eqref{eq_2d_problem}. Specifically, if its solution satisfies $\alpha_i\beta_i=0$, the corresponding rank-one component is identified as a pruning candidate; otherwise, it is retained. We refer to this optimization-induced zero--nonzero decision as an \emph{implicit thresholding rule} for LoRA rank allocation.

\subsection{LoRA Rank Allocation Based on Thresholding}

We now incorporate the proposed implicit thresholding rule into standard LoRA training. Importantly, the thresholding procedure does not replace the original optimizer used for fine-tuning. The LoRA factors are updated following standard training, while the proposed criterion is periodically invoked to identify redundant rank-one components.
We refer to the resulting method as $\ell_p$-LoRA, as its rank allocation is driven by $\ell_p$ regularization.
The detailed algorithm is presented in Algorithm~\ref{alg:pruning}.

\vspace{-0.5em}
\begin{algorithm}[h!]
	\caption{Rank Allocation via Thresholding ($\ell_p$-LoRA)}
	\label{alg:pruning}
	\begin{algorithmic}[1]
		\REQUIRE LoRA factors $\bA,\bB$, regularization parameter $\lambda$,
		curvature parameter $M$
		\FOR{each training iteration $k$}
		\STATE Update $\bA,\bB$ using an optimizer (e.g., AdamW)
		\STATE Aggregate gradients $\overline{\bG}_{A}$ and $\overline{\bG}_{B}$ as approximations to $\nabla_{\bA}\mL_k$ and $\nabla_{\bB}\mL_k$
		\IF{$k$ is a pruning step}
		\STATE $\widetilde{\bA}
		\leftarrow
		\bA-M^{-1}\overline{\bG}_{A}$, $\widetilde{\bB}
		\leftarrow
		\bB-M^{-1}\overline{\bG}_{B}$
		\FOR{$i=1,\ldots,r$}
		\STATE Solve~\eqref{eq_2d_problem} for
		$(\alpha_i^\star,\beta_i^\star)$
		\IF{$\alpha_i^\star\beta_i^\star=0$}
		\STATE Prune the $i$-th rank-one component
		\ENDIF
		\ENDFOR
		\ENDIF
		\ENDFOR
	\end{algorithmic}
\end{algorithm}

\vspace{-2.5em}
\section{Experiment}
\label{sec:typestyle}
\vspace{-0.5em}
\subsection{Natural Language Understanding}
We first evaluate the proposed method on natural language understanding tasks using DeBERTaV3-base~\cite{he2023debertav}. Experiments are conducted on four tasks from the GLUE benchmark~\cite{wang2018glue}, including CoLA, RTE, MRPC, and STS-B. Following the standard evaluation protocol, we report Matthews correlation for CoLA, accuracy for RTE, MRPC, and Pearson correlation for STS-B. We compare our method with LoRA~\cite{hu2022lora}, AdaLoRA~\cite{zhang2023adaptive}, and IGU-LoRA~\cite{cui2026igulora} under the same training schedule and a final rank budget equal to half of initial rank. For our method, we set $p=0.5$ and $M=10$, with the remaining pruning hyperparameters kept fixed across all tasks.
\vspace{-1em}
\begin{table}[h!]
\centering
\caption{Results on NLU tasks from the GLUE benchmark using DeBERTaV3-base. 
Higher values indicate better performance.
Results are reported as the mean over five random seeds, with the empirical standard deviation shown in the subscript. The best result is shown in bold.}
\label{tab:NLU_results}
\small
\setlength{\tabcolsep}{2.8pt}
\begin{tabular}{lcccc}
\hline
Method & CoLA & MRPC & RTE & STS-B \\
\hline

LoRA
& $69.10_{\pm0.28}$
& $90.10_{\pm0.51}$
& $87.40_{\pm0.85}$
& $91.82_{\pm0.12}$ \\

AdaLoRA
& $\textbf{70.57}_{\pm0.56}$
& $90.24_{\pm0.53}$
& $87.81_{\pm0.93}$
& $\textbf{91.94}_{\pm0.16}$ \\

IGU-LoRA
& $69.98_{\pm0.99}$
& $89.37_{\pm0.41}$
& $88.22_{\pm0.65}$
& $91.78_{\pm0.14}$ \\

$\ell_p$-LoRA
& $69.61_{\pm0.95}$
& $\textbf{91.32}_{\pm0.66}$
& $\textbf{88.45}_{\pm0.92}$
& $91.36_{\pm0.10}$ \\

\hline
\end{tabular}
\end{table}

Table~\ref{tab:NLU_results} summarizes the results. $\ell_p$-LoRA achieves the best performance on two out of the four tasks and remains competitive on the remaining tasks. In particular, the consistent performance across tasks with different evaluation metrics suggests that the proposed $\ell_p$-based implicit thresholding criterion can effectively identify redundant LoRA components while maintaining the adaptation capability of the model. 
Overall, these results demonstrate that rank allocation based on $\ell_p$ regularization provides a competitive alternative to existing fixed-rank and adaptive-rank LoRA methods.

\vspace{-1em}
\subsection{Question Answering}
We further evaluate the proposed method on question-answering tasks using Qwen2.5-7B~\cite{yang2024qwen2} to examine its effectiveness in a larger-scale language model and reasoning tasks. Experiments are conducted on BoolQ~\cite{clark2019boolq}, ARC-Easy~\cite{clark2018think}, OpenBookQA~\cite{mihaylov2018can}, and CommonsenseQA~\cite{talmor2019commonsenseqa}. All experiments are conducted under the same training setting and a
final rank budget equal to half of the initial rank, and report accuracy averaged over five random seeds. 

\vspace{-1em}
\begin{table}[h!]
\centering
\caption{Results on question answering tasks using Qwen2.5-7B. We report accuracy as the mean over five random seeds, with the empirical standard deviation shown in the subscript. 
The best result is shown in bold.}
\label{tab:qa_results}
\small
\setlength{\tabcolsep}{2.8pt}
\begin{tabular}{lcccc}
\hline
Method & BoolQ & ARC-Easy & OpenBookQA & CSQA \\
\hline

LoRA
& $89.25_{\pm0.12}$
& $92.39_{\pm0.83}$
& ${91.72}_{\pm0.83}$
& ${86.37}_{\pm0.53}$ \\

AdaLoRA
& ${89.42}_{\pm0.22}$
& $\mathbf{92.95}_{\pm0.23}$
& $\mathbf{92.20}_{\pm0.47}$
& $85.90_{\pm0.13}$ \\

IGU-LoRA
& $89.31_{\pm0.19}$
& ${92.74}_{\pm0.16}$
& $90.52_{\pm0.54}$
& $85.49_{\pm0.51}$ \\

$\ell_p$-LoRA
& $\mathbf{90.20}_{\pm0.38}$
& $92.17_{\pm0.49}$
& $91.12_{\pm1.04}$
& $\mathbf{86.39}_{\pm0.50}$ \\

\hline
\end{tabular}
\end{table}
As shown in Table~\ref{tab:qa_results}, our method achieves the best performance on two of the four tasks, including BoolQ and CommonsenseQA, while remaining competitive on ARC-Easy and OpenBookQA. In particular, the improvement on BoolQ is more pronounced, while the result on CommonsenseQA is comparable to the strongest baseline. These results indicate that the proposed thresholding-based rank-allocation criterion remains effective when applied to a larger language model across different question-answering tasks.

\vspace{-1em}
\section{Conclusion}
\vspace{-2mm}
In this work, we studied adaptive rank allocation for LoRA from a structured sparsity perspective. By imposing an $\ell_p$ penalty on the energy of rank-one LoRA components, we formulated rank allocation as an $\ell_p$-regularized optimization problem. We derived a tractable component-wise subproblem and reduced the coupled optimization to a two-dimensional scalar problem, leading to an implicit thresholding criterion for pruning redundant components while preserving standard LoRA training for the remaining ones. Experiments on natural language understanding and question-answering tasks demonstrate that the proposed method achieves competitive performance with existing fixed-rank and adaptive-rank LoRA methods under the same rank budget.

\bibliographystyle{IEEEbib}
\bibliography{strings,refs}

\end{document}